# Enhancing the Perception of Safety and Comfort during Physical Human-Robot Handshake Interactions by Integrating Flexible Elements into a Robotic Arm

Joel Hidalgo[1], Dennys Paillacho[1], Melissa Cobos[1] and Luigi Miranda[1]

***Abstract*— Safety and comfort in human-robot physical interactions are essential aspects in the development of social technologies, where natural gestures, such as handshakes, represent a challenge due to their direct physical contact. The implementation of series elastic actuators (SEA) to absorb impacts is proposed as a design strategy that favors safer interactions. This paper presents an experimental study aimed at evaluating how the incorporation of SEAs in robotic arms influences perceived safety and the interaction experience during handshaking. The design allows a direct comparison of the effect of rigidity versus the incorporation of elastic elements, in order to identify the advantages of SEAs in improving the physical safety and social acceptance of robotic systems in everyday contexts. The experiment was carried out with 10 volunteers (6 men and 4 women), who performed two interactions with each robotic arm: one with rigid joints and the other with flexible joints using SEA. During testing, objective data on end-effector trajectories were collected, as well as subjective information through a perception survey focused on safety, naturalness, and confidence during the handshake. The survey results show increased perceptions of safety and comfort with the SEA-equipped arm, supporting its potential to facilitate safer and more socially accepted human-robot interactions.**

## I. INTRODUCTION

Social robotics is based on the conception, design, and development of robotic systems capable of interacting smoothly with human beings. Unlike robots designed exclusively for operational functions in industrial settings or for carrying out repetitive tasks, social robots must have the ability to detect and interpret signals from human behavior, generate contextually appropriate responses, and integrate processes of understanding, logical reasoning, and movement execution according to specific demands [1], these capabilities allow them to dynamically adapt to the characteristics of human environments, facilitating their integration into social and collaborative contexts.

According to [2], a social robot is an autonomous agent that communicates and interacts with humans on an emotional level, as it follows patterns of social behavior and adapts to what it learns through its interactions. This type of interaction is based on visual and tactile perception as well as verbal communication [3]. Currently, the use of social robots has had a significant impact in various areas such as education, healthcare, entertainment, and research. Therefore, human–robot interaction plays a fundamental role in the different applications of social robots [4]. Physical contact, mainly touch in HRI, plays a crucial role since it can be used to convey information about a person's emotional state and thus create an emotional connection with the user during the interaction [5]. A natural physical interaction commonly used by humans is the handshake, which is one of the most common greetings and is often the first nonverbal interaction that occurs in a social context [6]. This gesture helps set the tone of communication, as different emotions can be conveyed through touch.

A predominant factor is safety, which represents the main priority in applications where physical interaction between humans and robots exists; therefore, the use of Series Elastic Actuators (SEA) has taken on a predominant role [7]. SEAs incorporate an elastic component between the gear motor and the load, achieving the decoupling of inertia and the nonlinear frictions of the transmission system from the load, which helps protect the actuator's internal components against impacts or improperly applied external forces. SEAs represent a superior alternative to rigid actuators in human–robot interaction contexts, providing advantages such as high compliance, reduced output impedance, greater tolerance to external impacts, and improved safety during contact with humans. These properties make them particularly suitable for advanced applications in robotics, such as the use of exoskeletons, robotic manipulators, and mobile robots, where mechanical efficiency and the safety of the operating environment are critical [8]. In rehabilitation robots, it is essential to reduce the adverse effects caused by rigid actuators; therefore, the implementation of SEA elements enables a safer and more biomechanically compatible interaction with the human body [9].

This work proposes the implementation of SEAs in the joints of a social robot's arm to enhance the system's intrinsic mechanical compliance during direct physical interactions. Unlike previous approaches in the handshaking literature, which predominantly focus on software-based algorithmic adaptations [10] or the kinematic breakdown of trajectories [6], there is a gap in evaluating how physical hardware compliance directly alters the user's perception of social acceptance. The integration of SEA elements into the robot's design aims to bridge this gap by emulating more flexible movements and reducing unintended impact risks. However, given the limited sample size analyzed ($N$ = 10), this work is explicitly defined as a preliminary investigation and an exploratory study. The primary objective is not to assert definitive statistical generalizability, but rather to qualitatively and descriptively identify initial trends in the participants' perception of safety and comfort under both joint configurations, serving as an engineering baseline for the subsequent design of active control strategies.

[1]All authors are with ESPOL Polytechnic University, Guayaquil, Ecuador. {joezhida,dpaillac,mncobos,luidamir}@espol.edu.ec

This paper is organized as follows. Section 2 reviews previous works where elastic actuators have been implemented to improve safety in interaction and in the evaluation of the handshake. Section 3 describes the design and experimental procedure. Section 4 presents the results obtained. Section 5 discusses and analyzes the findings, and finally, Section 6 outlines the conclusions of the study along with future research directions.

## II. Related Works

### A. SEA Implementations

The use of SEAs has been widely studied in various fields of robotics, mainly in the development of exoskeletons and robotic arms, due to their ability to enhance safety and mechanical compliance in human–robot interaction. Below, some relevant works are presented that support the proposal of this study.

In the work of [11], the design of a four-degree-of-freedom assistive exoskeleton aimed at handling heavy loads in work environments is proposed. This system incorporates SEAs in three of its joints, with the goal of reducing the risk of injuries in workers by providing mechanical support and improving ergonomics during object lifting. Complementarily, [12] employed SEAs in a three-DOF exoskeleton designed for shoulder rehabilitation. In that study, different configurations of linear tension springs arranged in disk and cylinder forms were explored, generating a nonlinear torsional behavior that allowed for a comparison of the efficiency of both configurations in terms of flexibility and performance.

In the context of collaborative robotics, [13] incorporated SEAs into industrial robotic systems in order to add a degree of passive mechanical compliance. This integration made it possible to improve cooperation between humans and robots by reducing the rigidity of physical interaction. In addition, different designs of elastic elements— including disc springs, helical springs, and torsion bars—were evaluated, showing that the choice of spring directly influences the impact absorption capacity and the dynamic behavior of the actuator.

Several studies have explored the potential of SEAs in robotic arms focused on safe interaction with humans. For example, [14] conducted a comparative analysis to determine how different types of actuators affect safety during a collision between a robotic arm and a person's head. The results showed that the use of SEAs significantly reduces the maximum impact force compared to conventional actuators, reinforcing their relevance in physical contact scenarios. Similarly, [15] implemented SEAs in two-DOF robotic arms, employing a fuzzy PI control system to mitigate the effects of external disturbances on the manipulated load. This approach demonstrated improvements both in system stability and in protection against unexpected forces.

Taken together, these works highlight the importance of using SEAs as a strategy to enhance safety, improve collaboration, and facilitate physical interaction between robots and humans. While most research has focused on exoskeletons and industrial applications, there are still opportunities to explore their incorporation in social and natural interaction contexts, such as the handshake gesture addressed in this work.

### B. Handshaking

A natural social interaction between humans is the gesture of shaking hands, either when meeting someone or when greeting before a meeting [16]. Therefore, in human–robot interaction, the handshake gesture acquires particular importance in social contexts, as it represents a symbolic action of trust and approach. Several studies have focused on examining human–human interaction as a preliminary step to designing the human–robot handshake [17] [10]. The study in [17] characterizes the handshake in terms of duration, grip strength, and frequency, while in [10] the movement is divided into three stages: reaching, handshaking, and hand withdrawal. Each of these stages has different particularities that can be expressed through mathematical formulations. On the other hand, several studies have addressed the generation of handshake movements in robots and their subsequent evaluation with users. For instance, [18] developed a kinematic controller for executing the handshake. More recently, [19] employed the Active Preference-Based Reward Learning (APReL) algorithm to optimize key parameters of the gesture, such as amplitude, frequency, and stiffness in the arm of a quadruped robot. To this end, experimental tests were conducted in which controlled variations of these parameters were introduced, and afterward, the individual preferences of the participants during the interaction were evaluated.

## III. Methods

This section describes the robotic platform used for the experiments, including the mechanical configuration of the robotic arms, the design of the incorporated elastic elements, the kinematic chain considered in the modeling, and the setup of the test bench for evaluating the handshake gesture. Furthermore, the methodology for collecting the trajectories during the handshake with participants is detailed.

For our analysis, the social robot Walter was used, a 1.3-meter-tall platform designed for research purposes in human–robot interaction. Walter has three degrees of freedom (DOF) in each upper limb, with dimensions that follow anthropometric ratios proportional to the size of the robot. The arm length is 19 cm, while the forearm length is 21 cm, which allows for an approximate emulation of the basic movements of a human arm in social interaction tasks, as shown in Figure 1.

The robotic arm considered in this study was modeled as an open kinematic chain with three degrees of freedom: shoulder rotation (vertical axis), shoulder elevation (frontal axis), and elbow flexion. This configuration allows for the description of the main movements required for the handshake gesture. The modeling was carried out using Denavit–Hartenberg (D-H) parameters, defining the relationship between the links and joints to obtain the position and orientation of the end effector. To enhance physical safety during interaction, a SEA was incorporated into one of the arms. For the fabrication of the elastic element, 3D printing with TPU (thermoplastic polyurethane) material was used due to its flexibility and impact absorption properties, which help mitigate the forces transmitted during contact. The geometric design of the spring was carried out considering compatibility with the arm joint and ease of integration into the platform. Figure 2 shows the manufactured model of the elastic element.

For the experimental validation, a test bench was built to evaluate the handshake interaction with people under controlled conditions. Two right robotic arms were installed on this bench, positioned at a height of 90 cm (corresponding to the mounting height on the Walter robot). Both arms have the same kinematic configuration and dimensions, differing only in the type of joints: one uses a conventional rigid system, while the other incorporates the flexible design with SEA. This setup allows for a comparison of behavior and perceived safety between a rigid arm and one with elastic elements. Figure 4 shows the configuration of the test bench.

### A. Experimental procedures

The present study aimed to analyze the differences in human-robot interaction during the execution of the handshake gesture, using two robotic arm configurations with different characteristics. Arm A was designed with rigid joints, while Arm B used flexible joints.

Ten participants (6 men and 4 women) participated in the experiment after receiving informed consent. Relevant anthropometric parameters, such as height and arm length (as shown in Table I), were recorded to explore the potential influence of participants' anthropometric characteristics on the observed interaction trajectories.

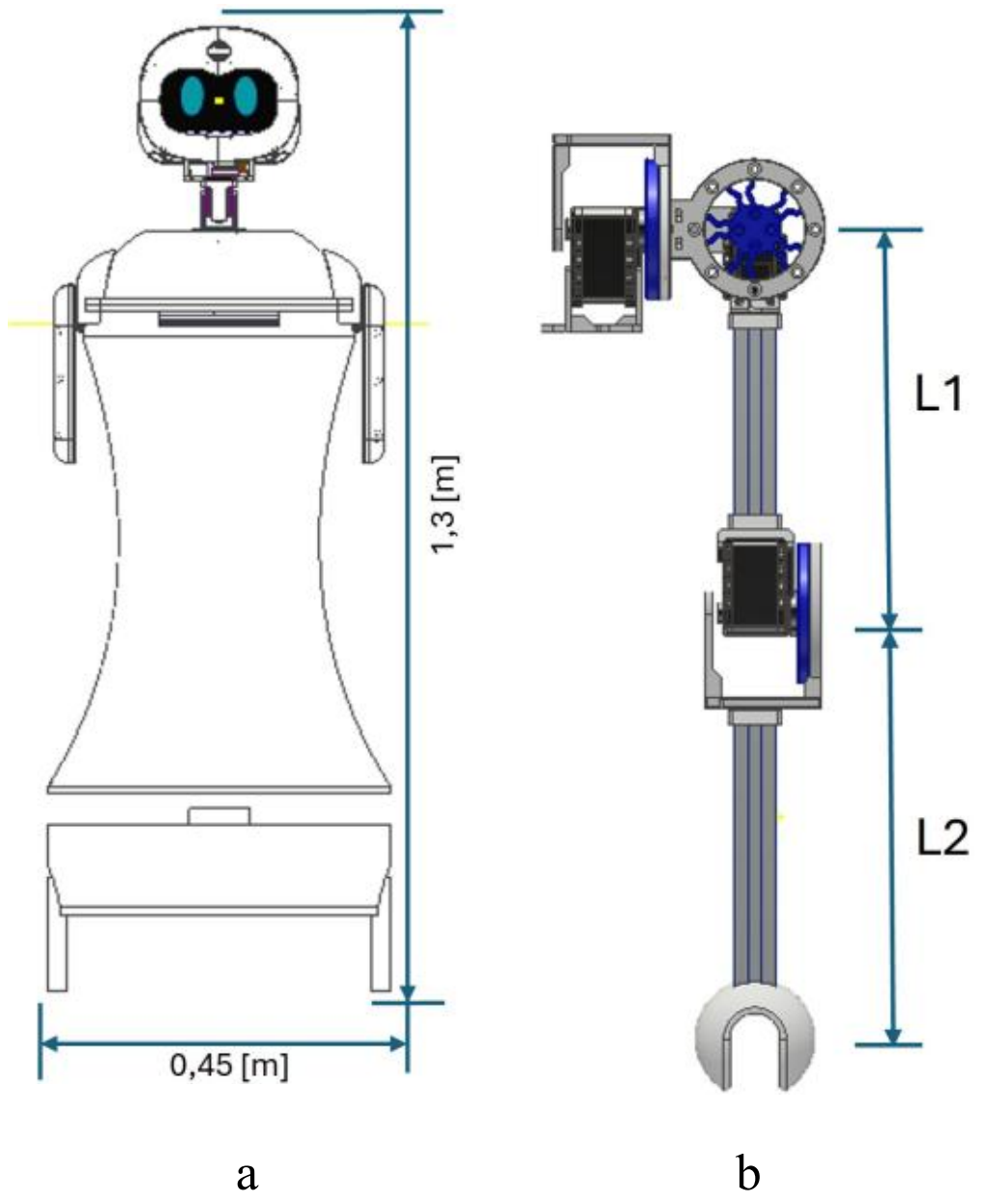

Fig. 1. a) Robot, b) Robot arm: L1 is 19 cm and L2 is 21 cm

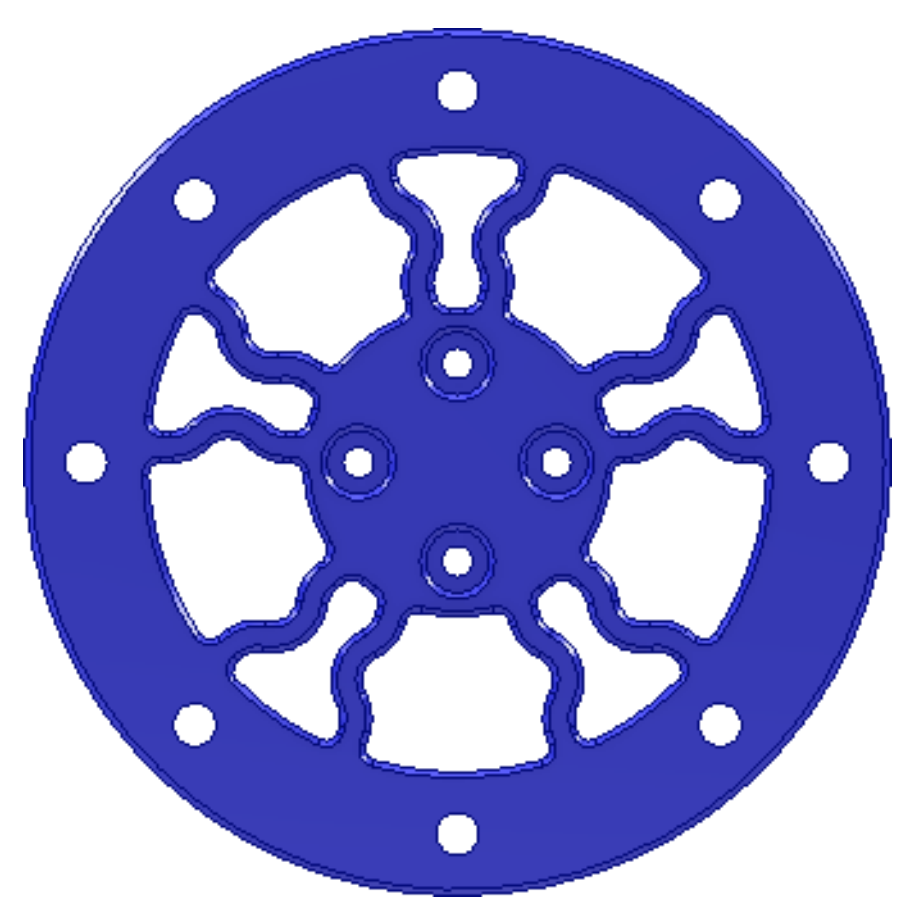

Fig. 2. Elastic element used in SEA

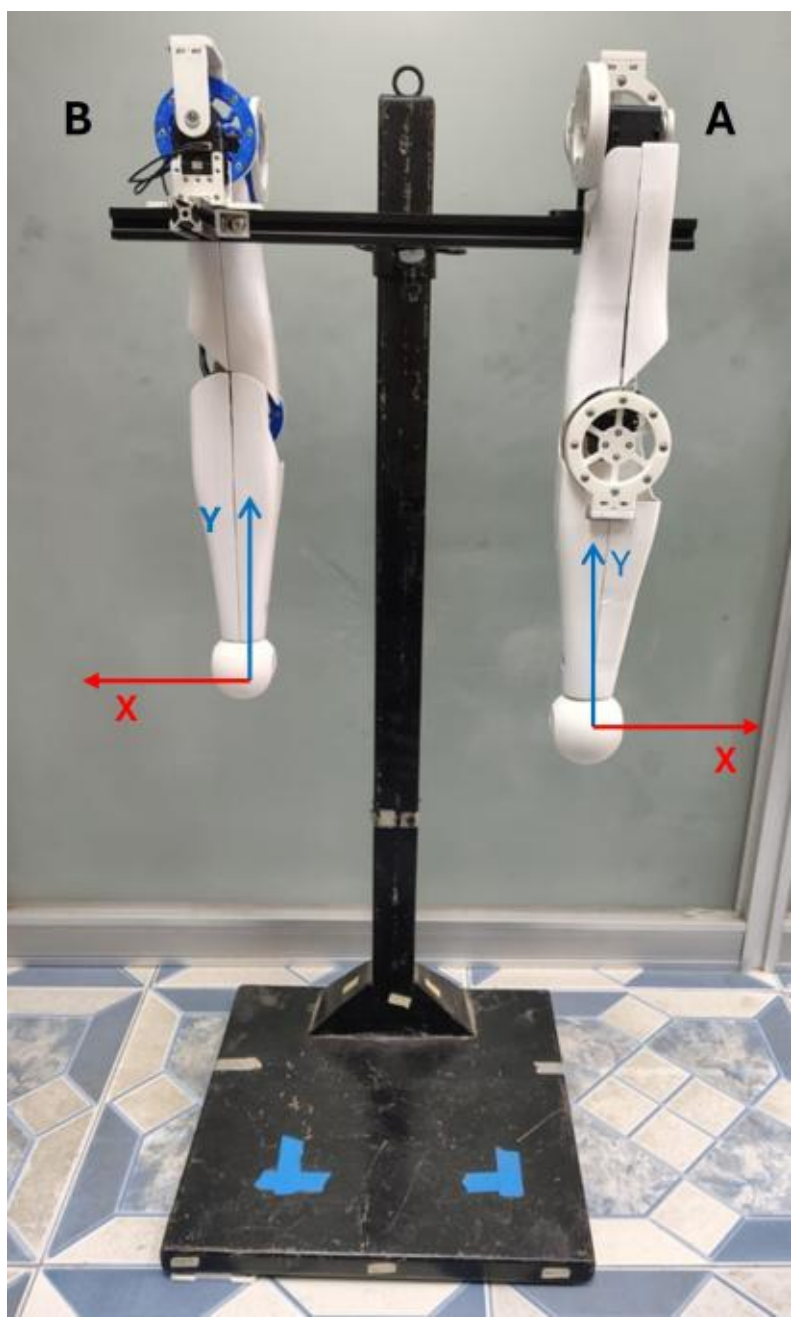

Fig. 3. Test bench. Arm A with rigid joints. Arm B with flexible joints

Each participant performed two trials with each arm, in which they had to execute the handshake gesture under controlled conditions. A total of 20 samples were collected between the two configurations. During each interaction, joint trajectories and contact time were recorded. Finally, each volunteer completed a perception survey.

## IV. Results

### A. Human-robot handshaking characterization

This study focused on a human–robot greeting interaction in which the robotic arm operated in a passive mode, allowing participants to guide the motion throughout the

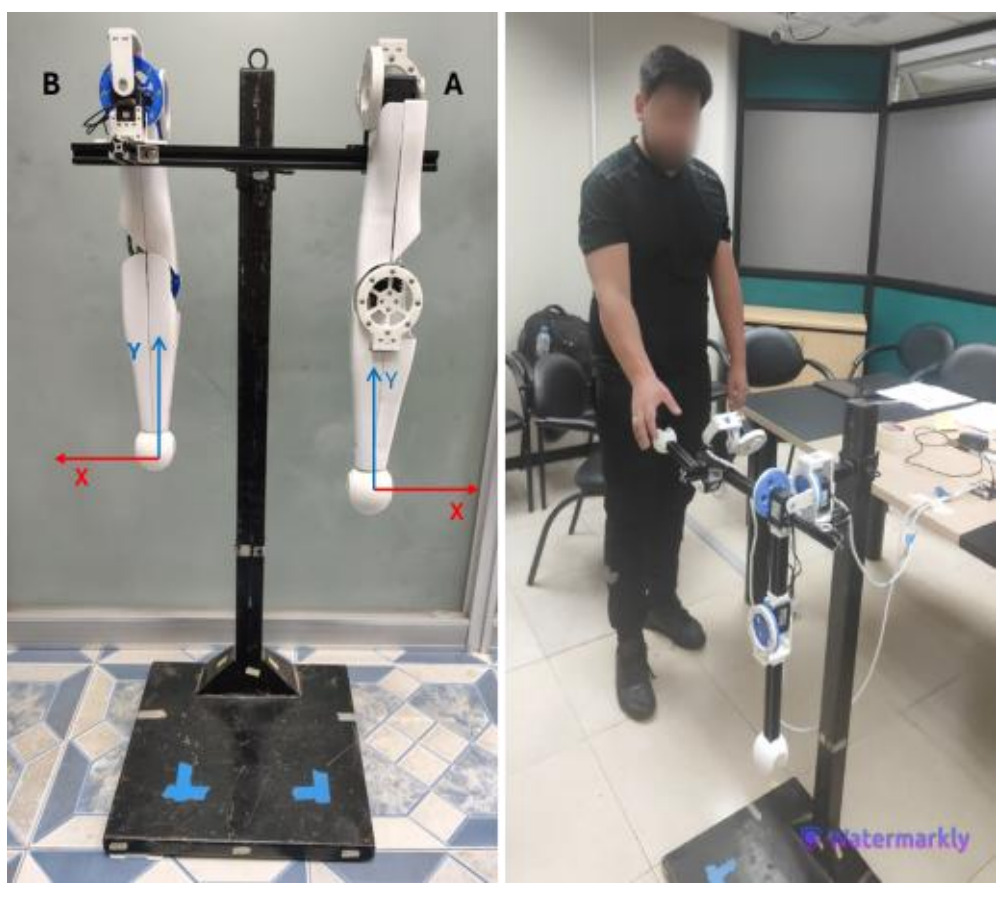

a b

Fig. 4. a) Test bench. Arms with flexible and rigid joints b) human-robot handshaking test

TABLE I

CHARACTERISTICS OF PARTICIPANTS

| Person | Gender | Height (m) | Arm length (m) |
|---|---|---|---|
| 1 | w | 1.58 | 0.68 |
| 2 | w | 1.57 | 0.70 |
| 3 | m | 1.90 | 0.85 |
| 4 | m | 1.80 | 0.81 |
| 5 | w | 1.59 | 0.70 |
| 6 | m | 1.77 | 0.77 |
| 7 | m | 1.77 | 0.74 |
| 8 | w | 1.64 | 0.66 |
| 9 | m | 1.70 | 0.78 |
| 10 | m | 1.80 | 0.81 |

handshake. In this configuration, the actuators remained unpowered, and no active control strategy, impedance regulation, or trajectory tracking was implemented. Consequently, all arm movements and oscillatory responses observed during the interaction were generated exclusively by the forces applied by the participant. This experimental design was intentionally adopted to isolate the influence of the mechanical characteristics of the robotic arm, specifically the presence or absence of SEAs, on user perceptions of safety and comfort while minimizing the influence of software-based compliance or control-related factors.

Before each trial, the robotic arm was actively positioned at a predefined initial location in the Cartesian workspace (x,y). Once the participant approached and established contact with the robot's hand, power to the actuators was disabled, placing the arm in a fully passive state. From that moment onward, the participant freely guided the handshake motion, and data acquisition was initiated to record the resulting trajectories and arm response during the interaction.

Data were recorded from the moment the human and robot hands made contact. During the experiment, the joint angles of the servomotors were stored, and using forward kinematics, the x,y coordinates corresponding to the end of the robotic arm were calculated, taking as reference the original position defined by the arm's resting state. It should be noted that this analysis was performed specifically on the arm with rigid joints, in order to characterize its behavior during the interaction.

From the graphs obtained during the experiment, particularities were identified that allow criteria to be established to characterize how a person executes the greeting gesture when interacting with a robotic system. In Figure 5 it can be observed that images **c, d, e, h** and **i** present two oscillations, while images **b, f** and **j** show three oscillations; in contrast, image a does not show these oscillations. It should be noted that this type of behavior is also found in greetings between humans, as reported in the literature [10].

### B. Surveys Results

To evaluate the proposed metrics, a survey was administered to participants after they had interacted with both arms. The results are presented in Figure 6, organized by each analysis criterion.

*a) Stiffness:* In the case of arm A (stiff joints), 40% of participants agreed that stiffness was perceived, whereas in arm B (with flexible joints), this percentage was 30%. However, in both cases, 50% of respondents disagreed with the statement that the arms did not feel stiff, indicating that the perception of stiffness was shared to some degree in both configurations.

*b) Comfort:* Arm B showed a descriptive improvement in user ratings, with 90% of participants agreeing or strongly agreeing that the greeting was perceived as more comfortable. In contrast, Arm A received 60% positive responses, 10% neutral, and 30% disagreed, indicating lower acceptance in terms of comfort.

*c) Naturalness:* In both arms, 50% of participants remained neutral regarding the naturalness of the greeting. However, 20% disagreed with Arm B, while Arm A received 40% disagreement. This suggests that, although the majority did not identify significant differences, Arm B was perceived as more natural than Arm A.

*d) Safety:* Regarding the feeling of safety, more marked differences were observed. Arm B achieved 50% agreement, compared to 30% for Arm A. Furthermore, only 10% of participants disagreed with the safety of Arm B, while this figure rose to 30% for Arm A. In both cases, 20% of respondents remained neutral and another 20% strongly agreed.

Overall, the results show that arm B, which incorporates serial elastic elements, was rated higher in terms of comfort and safety, while perceptions of rigidity and naturalness showed more balanced trends between the two arms. These findings suggest that the inclusion of flexible elements can contribute to a more comfortable and safe interaction experience, key aspects in the development of social robots geared toward physical interaction with people.

## V. DISCUSSION

The trajectories presented in Fig. 5 suggest that participant anthropometry was not a dominant factor in determining the observed motion patterns during the handshake interaction. Although it was expected that taller participants would generate greater vertical displacements and larger amplitudes along the y-axis, owing to differences in arm length and preferred handshake posture, no consistent relationship between participant height and the recorded trajectories was identified.

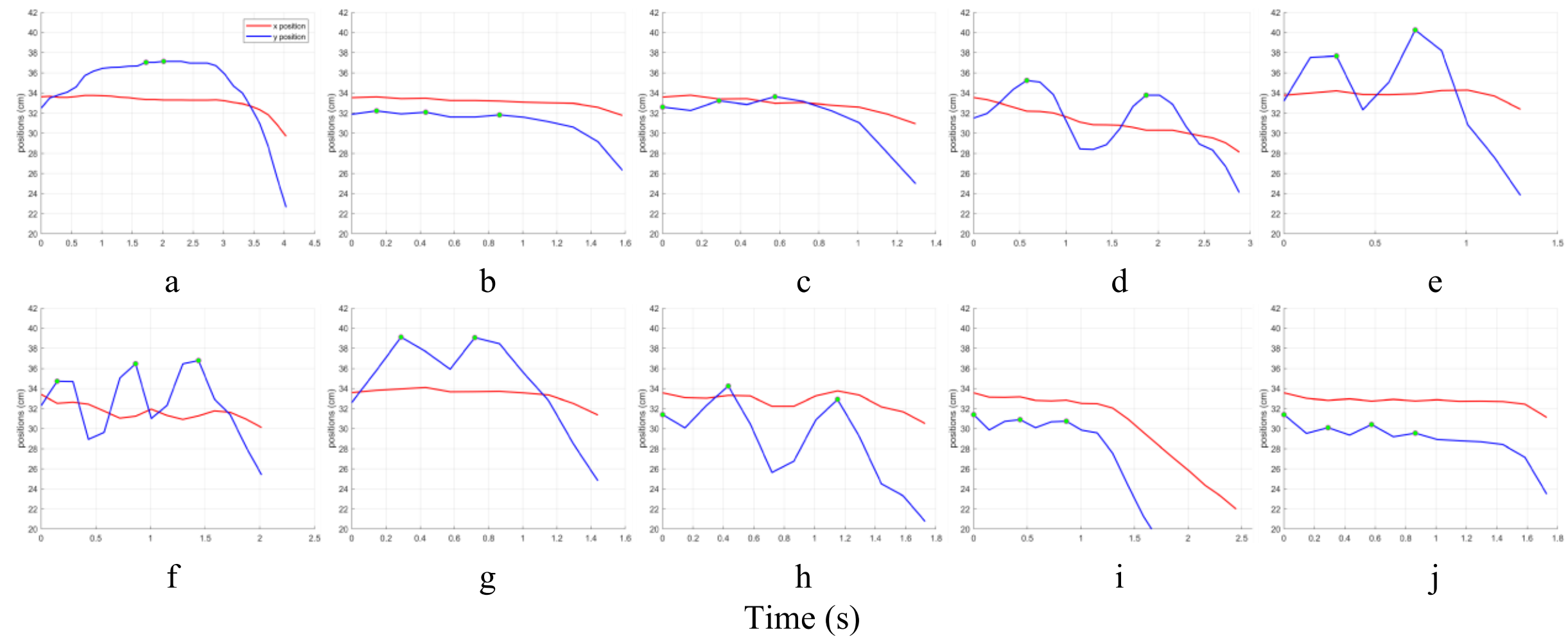

Fig. 5. Position vs time graphics of the robot hand during handshaking

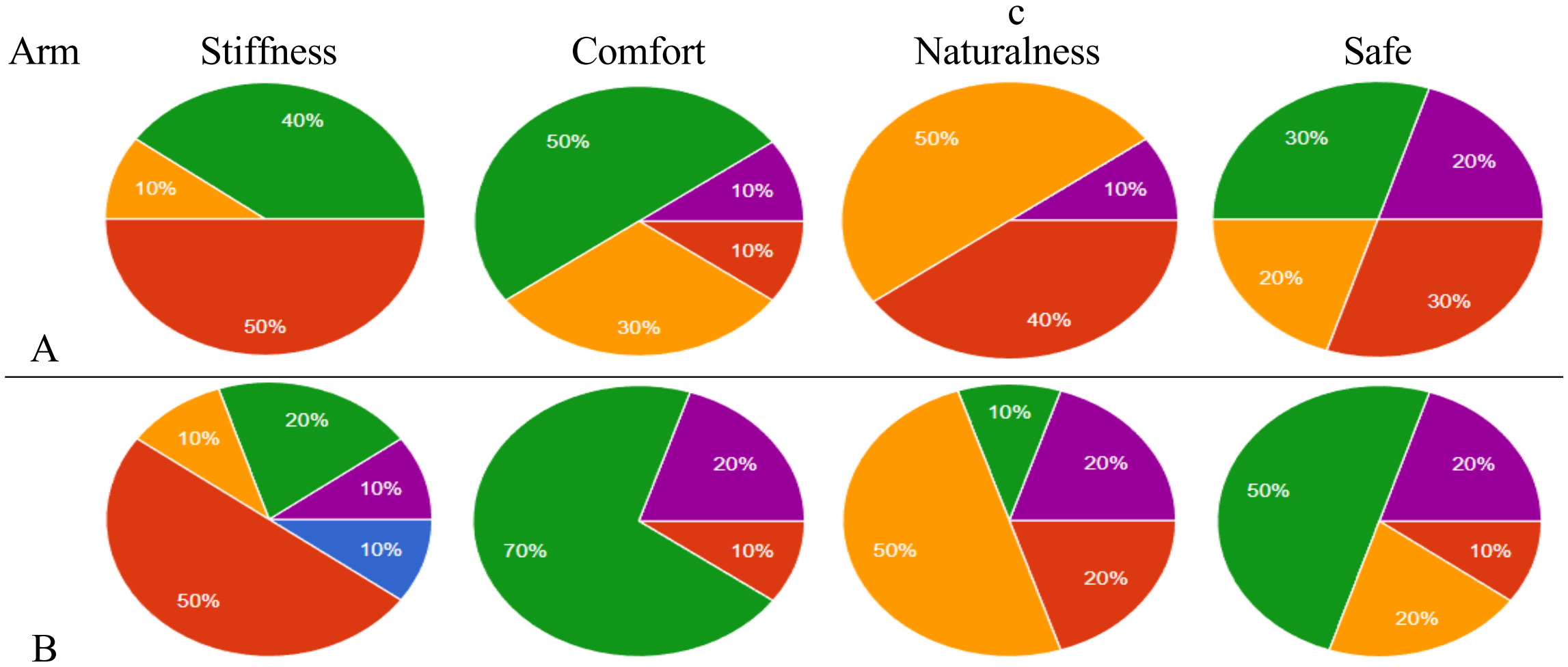

Fig. 6. Result of the perception survey of the criteria analyzed

One possible explanation is that the experimental setup constrained the interaction around a predefined initial contact position, reducing the influence of individual anthropometric differences on the resulting motion. Furthermore, participants appeared to adapt their posture and arm configuration to comfortably reach the robot, leading to more homogeneous interaction trajectories than initially anticipated.

An additional observation was that several participants did not perform the characteristic oscillatory motion typically associated with a human handshake. Instead, some interactions consisted mainly of establishing contact and executing limited arm movement. This behavior may be related to the passive nature of the robotic system. Unlike human-human handshakes, where mutual force exchange and haptic feedback contribute to movement synchronization [10], the robot did not actively respond or generate interaction forces. As a result, participants may have perceived the interaction as less reciprocal, reducing the tendency to perform a complete handshake motion.

On the other hand, the survey results suggest a preference for the SEA-based robotic arm, indicating that actuator compliance may enhance the user experience during physical interaction. Compared to the rigid configuration, the SEA arm achieved higher ratings in perceived comfort (90% versus 60%) and safety (50% versus 30%), supporting the initial hypothesis that compliant actuation contributes to more favorable human-robot interactions. These findings are consistent with previous studies on exoskeletons and robotic manipulators [8][9], which reported improvements in interaction safety and biomechanical compatibility through the use of series elastic actuators.

Regarding the naturalness of the gesture, the results were less conclusive: most participants adopted a neutral position, although a slight preference was observed for the flexible arm compared to the rigid one. This result suggests that, while flexibility contributes to the perception of a more human-like movement, the naturalness of the greeting depends on additional factors, such as timing, gesture variability, and adaptation to the user's individual characteristics—aspects previously discussed in works such as [17] [10].

The persistence of perceptions of rigidity in both configurations reveals that simply incorporating SEA is not sufficient to faithfully reproduce the dynamics of the human greeting. Therefore, there is a need to integrate more advanced control strategies, such as adaptive impedance control or reinforcement learning-based approaches, that allow for real-time adjustment of force and trajectory based on the user's response.

From an applied perspective, the findings have relevant implications for the design of social robots in educational, therapeutic, and collaborative contexts. The perception of safety and comfort not only increases technological acceptance but also fosters users' willingness to engage in meaningful physical interactions with a robot, a crucial aspect for the consolidation of social robotics in everyday life.

Finally, the small sample size (10 participants) is acknowledged as a major limitation, limiting the generalizability of the results. However, the study constitutes an initial contribution that validates the relevance of exploring SEA in social gestures such as greeting.

## VI. Conclusions and Future Work

This preliminary study investigated the impact of incorporating series elastic actuators (SEAs) into a robotic arm designed for social interaction through the handshake gesture. The results suggest that SEA-based configurations are perceived as more comfortable and safer than rigid alternatives, indicating that mechanical compliance may be a promising design strategy for improving the social acceptance of robots in collaborative and everyday environments.

Differences in perceived naturalness were less pronounced, although a slight preference for the flexible arm was observed. These findings support the hypothesis that mechanical flexibility can contribute to more human-like and trustworthy interactions. However, the persistence of rigidity-related perceptions and the generally neutral ratings of naturalness suggest that elastic elements alone may not be sufficient to fully reproduce the dynamics of a human handshake.

This observation highlights the potential need for additional mechanisms, particularly adaptive control strategies capable of regulating force, stiffness, and movement synchronization in real time. Such approaches could enhance the realism and responsiveness of physical human–robot interactions, although their effectiveness remains to be validated experimentally.

Overall, these results provide initial insight into the role of mechanical compliance in human–robot interaction and establish a foundation for future research. Future studies should include larger and more diverse participant groups, incorporate dedicated sensing technologies, and investigate AI-based motion generation and adaptive control approaches. Evaluations in broader social contexts may further clarify how mechanical compliance, sensing, and control contribute to safe, comfortable, and socially meaningful human–robot interactions.